\documentclass[11pt]{article}

\usepackage[utf8]{inputenc}
\usepackage[T1]{fontenc}
\usepackage{lmodern}
\usepackage{microtype}
\usepackage{booktabs}
\usepackage{amsmath}
\usepackage{graphicx}
\usepackage{xcolor}
\usepackage{hyperref}
\hypersetup{colorlinks=true, linkcolor=blue!50!black, citecolor=blue!50!black, urlcolor=blue!50!black}
\usepackage[round]{natbib}
\usepackage[margin=1in]{geometry}

\title{\textbf{Catching One in Five:\\ LLM-as-Judge Blind Spots in Production Multi-Turn Transaction Agents}}
\author{
  \textbf{Sawyer Zhang}\thanks{Corresponding author: \texttt{sawyer.zhang@lumivate.io}; ORCID \href{https://orcid.org/0009-0001-7736-1774}{0009-0001-7736-1774}} \quad
  \textbf{Alexander Wang} \quad \textbf{Sophie Lei}\\[4pt]
  \small Lumivate (Lumi)
}
\date{June 8, 2026}

\begin{document}
\maketitle

\begin{abstract}
LLM-as-judge has become the default instrument for evaluating conversational
agents, yet its reliability is almost always reported as \emph{agreement with
human ratings}, not as \emph{recall of real defects}. We study a deployed,
multi-turn food-and-beverage ordering agent and measure how many genuine
quality problems its built-in LLM judge actually catches, using exhaustive
human transcript review as ground truth. Across three review batches the
judge surfaces well under a quarter of human-confirmed systematic
problems---2 of 9 patterns ($22\%$) in one batch, and its operational gate
flagged \emph{zero} of 100 rounds in a batch where human review confirmed 23
distinct defects and 7 new cross-cutting patterns. We contribute a \emph{blind-spot
taxonomy} showing the failure is structured rather than random: the judge
reliably catches \emph{turn-local} issues (a fabricated statistic, a wrong
language) but systematically misses \emph{cross-turn state} issues
(confirm-gate lockout, cart hallucination, escalation lockout, stale
referents). We trace a concrete mechanism: the judge's scoring rubric exposes
only three coarse axes (intent, brand-voice, personalization) and has no
category for the behavioural dimensions---state-tracking, guardrails,
recovery---where most confirmed defects cluster. The failure is one of routing,
not perception: across the corpus, 113 of 114 rounds whose raw judge note
describes a confirm-gate or cart-state defect are scored ``brand voice'', and
none reach an operational failure---the shipping gate is wired to hangs and hard
assertions, not to the rubric---so the $0\%$ is a routing-and-wiring failure, not
blindness. The consequence for prevalence estimation
is sharp: when the gate's apparent defect rate is zero the Rogan--Gladen
correction degenerates---no apparent-prevalence signal can recover the true rate
at all, a stronger statement than any fixed multiplier---while where the gate
reports a nonzero rate the same estimator implies a $3$--$6\times$
(order-of-magnitude) undercount under our measured sensitivity. We argue that
for production multi-turn agents automated judging is a regression floor, not a
substitute for human review.
\end{abstract}

\section{Introduction}
LLM-as-judge~\citep{zheng2023judging,liu2023geval} is now standard for
evaluating open-ended and agentic systems: it scales, it is cheap, and on
single-turn preference tasks it correlates well with human raters. A large
literature studies its \emph{biases}---position, verbosity, self-preference
\citep{ye2024justice,shi2024position,wu2024selfpref}---and its
\emph{agreement} with humans \citep{gu2024survey,judgesverdict2025}. Far less
is known about a property that matters more for shipping software: when an
LLM judge runs as the automated quality gate on a production agent's live
traffic, \emph{what fraction of the real problems does it actually catch?}

We answer this for one concrete, deployed system: a per-tenant, multi-turn
ordering agent for South-East-Asian food-and-beverage brands, evaluated on a
randomized multi-turn corpus with exhaustive human transcript review as
ground truth. Our findings are uncomfortable and, we argue, generalizable in
shape if not in exact value:
\begin{itemize}
\item \textbf{Low recall, not just low agreement.} The built-in judge catches
well under a quarter of human-confirmed systematic problems. In one batch it
auto-captured 2 of 9 human-identified patterns ($22\%$); in a 100-round batch
it flagged \emph{zero} turns as failures while human review confirmed 23
distinct ledger defects and promoted 7 new patterns.
\item \textbf{The blind spot is structured.} Recall degrades monotonically
with how ``cross-turn'' a problem is. Turn-local, surface-checkable issues
(fabricated statistics, wrong-language replies) are caught; issues that
require reasoning over the conversation arc (confirm-gate state, cart
contents, escalation flags, last-proposal referents) are missed.
\item \textbf{A mechanism, not a mystery.} The judge's own rubric has three
axes (intent / brand-voice / personalization). It has \emph{no} bucket for
state-tracking, guardrails, or recovery---the dimensions that hold most
confirmed defects---so confirm-gate violations get filed as ``brand voice''
or slip through.
\end{itemize}
We position this against, and distinguish it from, the two closest lines of
work---single-turn error-detection studies that already report low judge
recall~\citep{kamoi2024realmistake}, and the one benchmark that judges the
\emph{judges of agents}~\citep{agentrewardbench2025}---in \S\ref{sec:related}.

\section{Related Work}
\label{sec:related}
\paragraph{LLM-as-judge: bias and agreement.} The canonical
formulation~\citep{zheng2023judging} and CoT variants~\citep{liu2023geval}
report reliability as correlation/agreement with human preference. A rich
follow-up quantifies systematic
\emph{biases}~\citep{ye2024justice,shi2024position,wu2024selfpref} and revisits
agreement with stricter statistics~\citep{judgesverdict2025,gu2024survey}.
This line answers ``does the judge rank like a human?''---not ``what does the
judge miss?''.

\paragraph{Judge recall / false negatives.} Closest in spirit,
\citet{kamoi2024realmistake} show GPT-4/Claude detect real errors in LLM
responses at \emph{very low recall}, far below humans, and that
self-consistency does not help; \citet{jain2025beyond} report true-negative
rates below $25\%$ and note class imbalance inflates accuracy. Both are
\emph{single-turn, objective-error} settings. Our contribution is the
\emph{multi-turn, production transaction-agent} instantiation and, crucially,
a taxonomy explaining \emph{which} defect classes are missed.

\paragraph{Judging agent trajectories.}
\citet{agentrewardbench2025} is the genuine near-twin: the first benchmark
assessing LLM judges of agent trajectories (expert-reviewed web-agent runs),
finding no judge excels everywhere. It covers single-trajectory web agents
and does not report a production false-negative-rate-versus-human-review
headline, nor a turn-local-versus-cross-turn-state taxonomy---the two gaps we
fill.

\paragraph{Agent evaluation by outcome.} Mainstream agent
benchmarks~\citep{yao2024taubench,zhou2024webarena,liu2024agentbench,ma2024agentboard,survey2025multiturn}
score end-to-end task success. Notably \citet{yao2024taubench}, the closest
to our transaction setting, uses database-state goal-matching rather than an
LLM judge---an outcome oracle that, like our findings, sidesteps judge recall
but cannot diagnose \emph{where} behaviour broke.

\paragraph{Correcting noisy judges.} Using a calibrated noisy classifier to
recover true prevalence is the Rogan--Gladen
estimator~\citep{rogan1978}; its application to LLM judges is very
recent~\citep{chen2026noisy,noisyvalid2026}. We adopt this framing
(\S\ref{sec:rg}) to argue that a measured judge false-negative rate should
\emph{correct}, not merely accompany, reported quality.

\section{Setting: A Production Multi-Turn Ordering Agent}
\label{sec:setup}
The system under study is a per-tenant brand agent that takes food-and-drink
orders in natural language over multi-turn conversations: it searches a menu,
builds and mutates a cart, answers product questions, applies dietary
constraints, and places orders behind an explicit confirmation gate. The
evaluation tenant is a Singapore coffee brand with a bilingual
(English / Mandarin, with code-switching) customer base.

\paragraph{Corpus.} We use a randomized multi-turn corpus generated by a loop
driver that samples a scenario \emph{family} (e.g.\ \texttt{the\_usual},
\texttt{deep\_dietary}, \texttt{large\_group}, \texttt{cancel},
\texttt{slot\_conflict}), a locale (\texttt{en-SG}/\texttt{zh-SG}), an
identity mode (anonymous / verified-with-favourites), and a depth of 5--9
user turns. The batches analysed below span 34 scenario families, balanced
50/50 across locales, with conversation depths concentrated at T5--T9
(median latency $73$\,s, p95 $192$\,s per round).

\paragraph{The built-in judge.} During each loop round an LLM judge scores
the transcript against a rubric that emits three \emph{capability} axes---%
\texttt{intent}, \texttt{brand\_voice}, \texttt{personalization}---each a
pass/fail bit with a free-text reason, plus an optional free-text
\texttt{suspected\_bug} note. These axis scores feed an auto-export
\emph{regression ledger}: a case is recorded whenever an axis is scored $0$ and,
once it recurs, auto-promoted to a fixture. Separately, the loop emits the
single \emph{operational gate} verdict, \texttt{live\_passed}---the output that
actually ships and blocks. The two are not merely reported separately; they are
\emph{architecturally disconnected}. We verified in the loop driver that
\texttt{live\_passed} is \emph{not} a function of the three axes: it flips to
fail only on an operational failure (the agent raises an exception or returns no
response) or a hard per-turn assertion---never on a low quality score, however
severe. So the quality rubric and the gate that ships have no wiring between
them, and we measure them separately because they diverge sharply
(\S\ref{sec:notice}). ``Judge recall'' in our headline refers to the operational
gate---the output that ships---unless stated otherwise.

\paragraph{Ground truth.} Our ground truth is exhaustive human transcript
review: an analyst (with, in the largest batch, an adversarial multi-reviewer
workflow---one reviewer per family followed by a skeptic verifier prompted to
\emph{refute} each finding before synthesis) reads full turns plus tool calls
and records a \emph{pattern} only when a behaviour recurs across multiple
rounds, families, or locales. A pattern is the symptom of a \emph{layer}, not
a one-off. This yields a conservative denominator: single occurrences are
excluded.

\section{Method}
We compare, per batch, the set of human-confirmed problems against the set
the judge flagged. We report (i) \emph{pattern recall}: of the systematic
patterns human review confirmed, how many the judge's pipeline captured;
(ii) the \emph{distribution of confirmed problems over behavioural
dimensions} versus the judge's available rubric axes; and (iii) a
\emph{blind-spot taxonomy} keyed on whether a defect is decidable from a
single turn or requires the conversation arc. All numbers below are extracted
from the project's pattern log and auto-captured regression ledger; the
exact batch identifiers are retained for reproducibility.

\section{Results}

\subsection{Recall is well under a quarter}
\label{sec:recall}
\paragraph{Batch B1 (R166--R196, deep-flow).} Exhaustive human review of the
R166--R196 deep-flow batch surfaced 9 systematic patterns. The operational
gate auto-captured 2 of them---a \textbf{$22\%$ pattern recall} (Wilson $95\%$
CI $[6.3\%, 54.7\%]$; the interval is wide at $n{=}9$). The point of B1 is not
a precise rate but that the upper end of even a generous interval sits near
half. The $\approx\!18\%$ figure we quote elsewhere as a \emph{running}
estimate is the program-level pooled ratio---auto-captured patterns as a
fraction of human-confirmed patterns, pooled across loop batches rather than
computed on B1 alone---and it sits just below the B1 point because later
batches confirmed more patterns than the gate captured; we use it only as an
order-of-magnitude anchor, never as a precise rate.

\paragraph{Batch B2 (R166--R265, 100 deep rounds).} On the closed 100-round
batch the operational gate's failure rate (\texttt{live\_passed}) was
\textbf{$0\%$}: it surfaced no round as failing. Human review of the same
rounds confirmed \textbf{23 distinct defects} in the ledger and promoted
\textbf{7 new patterns} beyond B1. With zero of 23 confirmed defects surfaced,
the gate's recall on this batch is $0\%$ with a rule-of-three $95\%$ upper
bound of only $13\%$ (Wilson $[0\%,14.3\%]$); at the round level, $0/100$
rounds flagged gives an upper bound of $3.7\%$. On this batch the operational
gate's recall for the multi-turn defect classes involved is statistically
indistinguishable from zero. Crucially, this $0\%$ is \emph{not} the raw judge
falling silent---\S\ref{sec:notice} shows the raw judge frequently \emph{notices}
these defects in free text but the rubric vocabulary and a gate disconnected
from it prevent the notice from becoming a surfaced failure.

\paragraph{Batch B3 (R1--R160, post-fix, fresh pool).} An adversarial
human-in-the-loop review (one reviewer per family $\rightarrow$ refute-first
skeptic verifier $\rightarrow$ synthesis) raised \textbf{147 suspected issues},
of which \textbf{85 were adversarially confirmed} (22 explicitly refuted as
reviewer/judge artefacts), skewed to high severity---money, state, and
guardrail P0/P1 defects dominate the confirmed set. This batch established 14
further patterns (P17--P30), several of which were \emph{regressions} of
B1/B2 fixes---defects the automated gate again did not surface as failures.

Across batches (Table~\ref{tab:batches}), the consistent picture is an
operational gate that surfaces a small minority of the systematic, multi-turn
problems a human finds. We deliberately do \emph{not} collapse this to a single
clean number: the three batches enumerate different units (systematic patterns,
ledger-confirmed defects, rounds), so $22\%$ (B1), $\approx 0\%$ (B2), and the
$\approx 18\%$ pooled running estimate are not the same quantity---they bound
the same \emph{conclusion}, with every interval's upper edge well under half.
The one unit common to a batch is the round (B2: $0/100$); a fully turn-level
recall comparable across batches needs per-turn human labels and is part of the
calibration pass of \S\ref{sec:rg}.

\begin{table}[t]
\centering
\small
\begin{tabular}{lllr}
\toprule
\textbf{Batch} & \textbf{Scope} & \textbf{Judge recall} & \textbf{95\% CI} \\
\midrule
B1 (R166--R196) & 9 patterns      & $2/9 = 22\%$  & $[6.3, 54.7]$ \\
B2 (R166--R265) & 23 defects      & $0/23 = 0\%$  & $[0, 14.3]$ \\
B2 (round-level)& 100 rounds      & $0/100 = 0\%$ & $[0, 3.7]$ \\
program est.\   & rolling         & $\approx 18\%$ & --- \\
\bottomrule
\end{tabular}
\caption{Built-in judge recall against exhaustive human review, per batch
(Wilson intervals, in \%). Denominators differ by what each batch's human
review enumerated (systematic patterns / ledger-confirmed defects / rounds);
all bound the same conclusion---recall well under a quarter, upper edge under
half even at $n{=}9$.}
\label{tab:batches}
\end{table}

\subsection{The defect distribution the judge cannot see}
\begin{figure}[t]
\centering
\includegraphics[width=0.96\linewidth]{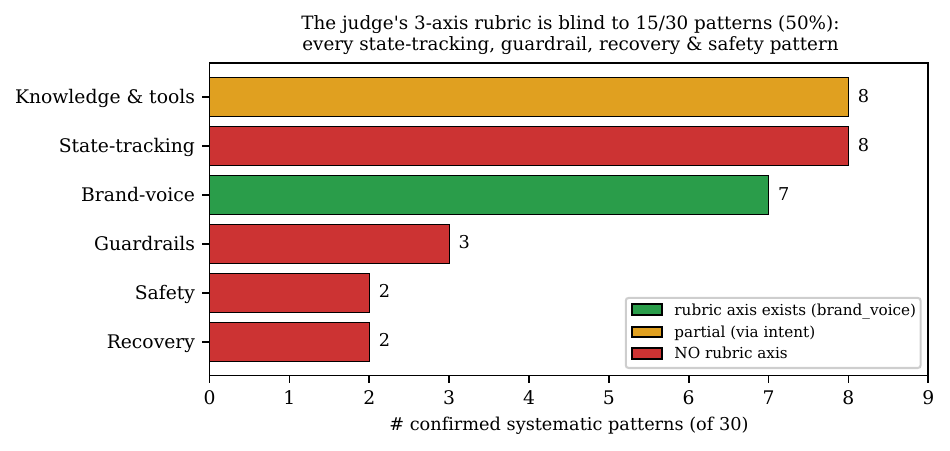}
\caption{The coverage gap, visualized. Each bar is a behavioural dimension's
share of the 30 confirmed patterns, colored by whether the judge's three-axis
rubric can name it: green = a dedicated axis (\texttt{brand\_voice}); amber =
partial (folded into \texttt{intent}); red = \emph{no} axis. Every red
dimension---state-tracking, guardrails, recovery, safety, $15/30$
patterns---is structurally invisible to the judge's scoring vocabulary.}
\label{fig:coverage}
\end{figure}

Table~\ref{tab:dims} and Figure~\ref{fig:coverage} map the 30 confirmed
patterns (P1--P30) onto five behavioural dimensions and contrast them with the
three axes the judge's rubric can emit. The mismatch is the mechanism: \textbf{15 of 30
confirmed patterns ($50\%$)---every state-tracking, guardrail, recovery, and
safety pattern (8/3/2/2)---fall into dimensions the judge's rubric has no
category for}, a clean $50/50$ split against the brand-voice and knowledge
patterns it can at least partially name. Inspection of the auto-captured
ledger confirms the leakage concretely: confirm-flow and order-completion
failures (guardrail/state defects---the agent mishandling an explicit
``confirm'') appear in the ledger tagged \texttt{brand\_voice} (e.g.\ the
\texttt{stacked\_mutation\_confirm} and \texttt{order} families).

\begin{table}[t]
\centering
\small
\begin{tabular}{lcl}
\toprule
\textbf{Human behavioural dimension} & \textbf{\# patterns (of 30)} & \textbf{Judge rubric axis?} \\
\midrule
State-tracking (cart, referents, quantity) & 8 & \textcolor{red}{none} \\
Knowledge \& tools                         & 8 & partial (via \texttt{intent}) \\
Brand-voice                                & 7 & \texttt{brand\_voice} \\
Guardrails (confirm-gate, budget)          & 3 & \textcolor{red}{none} \\
Recovery (escalation lockout, OOD)         & 2 & \textcolor{red}{none} \\
Safety (allergen, stock-out)               & 2 & \textcolor{red}{none} \\
\midrule
\textbf{Total}                             & \textbf{30} & 3 axes available \\
\bottomrule
\end{tabular}
\caption{Confirmed systematic patterns by behavioural dimension versus the
judge's available rubric axes. The judge can emit only
\texttt{intent}/\texttt{brand\_voice}/\texttt{personalization}; it has no
native category for state-tracking, guardrails, recovery, or safety---which
together hold $50\%$ of confirmed patterns. Patterns carrying more than one
dimension tag are counted under their primary dimension.}
\label{tab:dims}
\end{table}

The judge's own captured ledger (74 family$\times$capability entries) is
distributed \texttt{brand\_voice} 43, \texttt{intent} 28,
\texttt{personalization} 3---i.e.\ the judge funnels the multi-turn failures
it \emph{does} notice into ``brand voice'', the one rubric axis loose enough
to absorb them, which both mislabels the defect and prevents per-dimension
triage.

\subsection{What the judge notices but cannot surface}
\label{sec:notice}
The $0\%$ operational recall (\S\ref{sec:recall}) could mean two very different
things: the raw judge is \emph{blind} (never registers the defect), or the raw
judge \emph{registers but cannot route} it---it notices the problem, has no
rubric axis to file it under, and the shipping gate never consumes the
notice at all. These have opposite design implications, so we separate
them directly using the judge's raw per-axis scores and free-text
\texttt{suspected\_bug} notes, logged alongside the operational verdict. Over
the current 220-round corpus the split is unambiguous:
\begin{itemize}
\item The \textbf{operational gate} (\texttt{live\_passed}) reports
\textbf{$0/220$} failures---consistent with the per-batch recall above.
\item The \textbf{raw judge is far from silent}: it emits a free-text
\texttt{suspected\_bug} on \textbf{$125/220$} rounds ($57\%$).
\item Of those notes, \textbf{$114$ describe a confirm-gate or cart-state
defect}---precisely the cross-turn classes---and \textbf{$113$ of $114$
($99\%$) are scored under \texttt{brand\_voice}}, the only loose axis available.
\item \textbf{None of the $125$ notices became an operational failure}
($0/220$ \texttt{live\_passed} fails)---and not because a disposition step
weighed and dropped them, but because \texttt{live\_passed} is not wired to the
quality axes at all (it fails only on a hang or a hard assertion, \S\ref{sec:setup}).
\item \textbf{This is structural, not a one-run artifact.} On an independent
earlier corpus ($381$ rounds) the gate \emph{did} fail---$121$ times---but
every one was an operational hang (the agent returned no response), not a
quality catch; and of the $144$ rounds there where the agent \emph{responded}
and the judge \emph{emitted} a quality note, all $144$ ($100\%$) still passed
the gate. Pooling both corpora, $269$ responded-with-a-quality-note rounds
produced \textbf{zero} gate failures. The gate's blindness to quality is a
property of the wiring, not of any particular run.
\end{itemize}
Table~\ref{tab:routing} breaks the $125$ notices down by what the note
\emph{describes} (a transparent keyword heuristic over the judge's own free
text) against what the rubric \emph{scored} and whether the gate escalated.
Two rows are the mechanism. Notes that genuinely describe a tone/voice problem
\emph{should} score \texttt{brand\_voice}, and do---that is correct routing.
But the confirm-gate/guardrail and cart/state notes, which describe behavioural
defects with no matching axis, also collapse onto \texttt{brand\_voice}
($113/114$); and the final column is uniformly zero---not one of the $125$
notices, of any theme, survives to become a gate failure (the gate never reads them).
The misrouting and the dissolution are visible in the same table.

\begin{table}[t]
\centering
\small
\begin{tabular}{lccc}
\toprule
\textbf{What the note describes} & \textbf{Notes} & \textbf{Scored \texttt{brand\_voice}} & \textbf{Became gate failure} \\
\midrule
Confirm-gate / guardrail & 105 & 104 & 0 \\
Cart / state            &   9 &   9 & 0 \\
Brand-voice / tone      &   1 &   1 & 0 \\
Knowledge / tools       &   2 &   2 & 0 \\
Other / unclassified    &   8 &   5 & 0 \\
\midrule
\textbf{Total}          & \textbf{125} & \textbf{121} & \textbf{0} \\
\bottomrule
\end{tabular}
\caption{Raw-judge \texttt{suspected\_bug} notes ($125/220$ rounds), classified
by theme, against the rubric axis scored and the operational outcome. Themes
are assigned by a keyword heuristic over the free-text note (the script and the
per-round audit list are released); the load-bearing rows are robust to
classifier slop because the behavioural-vs-\texttt{brand\_voice} share is
near-saturated ($113/114$) and the final column is identically zero regardless
of theme. Reproduced on the current 220-round corpus by
\texttt{eval/experiments/p1\_judge\_routing.py}.}
\label{tab:routing}
\end{table}
So the blind spot is not (only) cognitive. The raw judge routinely \emph{detects}
the confirm-gate and state defects; what fails is \emph{recording and acting}:
the rubric has no state/guardrail axis, so a detected guardrail violation is
written as ``brand voice'' ($113/114$), and the shipping gate---which never
reads the axes---cannot act on it regardless. This is the mechanism of
\S\ref{sec:recall} pinned to data---a rubric-routing failure compounded by a
gate disconnected from the quality signal, not merely a perception failure---and
it sharpens the fix (\S\ref{sec:design}): giving the rubric the missing axes is
not cosmetic relabeling but the prerequisite for the notice the judge
\emph{already produces} to reach the gate at all.

\subsection{Blind-spot taxonomy: turn-local vs.\ cross-turn}
Table~\ref{tab:taxonomy} arranges representative confirmed patterns by how
much conversational context is needed to decide the defect. Judge recall
falls as the required context widens from a single turn to the full session
arc. Turn-local, surface-checkable defects (a tilde-prefixed fabricated
statistic; a Vietnamese reply in an English session) are caught or are at
least matchable by a regex; cross-turn-state defects (confirm-gate lockout,
cart hallucination, escalation lockout, stale last-proposal) require holding
and comparing state across turns---exactly what a per-turn-scored judge does
not do.

\begin{table}[t]
\centering
\small
\begin{tabular}{lllc}
\toprule
\textbf{Defect class (example pattern)} & \textbf{Dimension} & \textbf{Context needed} & \textbf{Judge catches?} \\
\midrule
Fabricated statistic ``$\sim$17\% of orders'' (P4) & brand-voice & single turn & medium \\
Wrong-language reply (P21)                          & brand-voice & single turn & medium \\
Hallucinated price/total (P12)                      & tools       & single turn$^\dagger$ & low \\
Cart hallucination / phantom add (P11, P22)         & state       & cross-turn  & low \\
Confirm-gate bypass / lockout (P1, P17)             & guardrails  & cross-turn  & very low \\
Escalation lockout after ``nevermind'' (P3)         & recovery    & session arc & $\approx$ none \\
Last-proposal referent break (P5, P25)              & state       & session arc & $\approx$ none \\
\bottomrule
\end{tabular}
\caption{Blind-spot taxonomy. Recall degrades as the context required to
decide the defect widens. $^\dagger$Price defects are nominally single-turn
but require reconciling the quoted number against tool-returned cart state,
which the judge does not do.}
\label{tab:taxonomy}
\end{table}

\subsection{Three case studies the judge passed}
\label{sec:cases}
The taxonomy is concrete in the data. Below are three real scenarios (verbatim
user turns from the corpus; the agent's defect is the documented pattern; the
judge's verdict is from the live run) where the judge emitted no failure---and,
where it tagged anything at all, tagged \texttt{brand\_voice}, a dimension that
does not describe the defect.

\medskip\noindent\fbox{\begin{minipage}{0.95\linewidth}\small
\textbf{Case 1 --- confirm-gate, tagged \texttt{brand\_voice} (P1/P17).}
User: \textit{``what's my cart and total? $\mid$ anything change? if it's all
good, confirm the order.''} The defect class is a guardrail/state contract:
either an order is placed without an unambiguous confirm, or (post-fix) an
explicit confirm fails to place it. Deciding it requires comparing the
confirm-utterance against the prior proposal and the resulting tool calls---two
turns of state. The judge logged this family under \texttt{brand\_voice}.
\end{minipage}}

\medskip\noindent\fbox{\begin{minipage}{0.95\linewidth}\small
\textbf{Case 2 --- mid-session allergen + confirm, tagged \texttt{brand\_voice}
(P15/safety).} User: \textit{``what's in my cart? $\mid$ actually I just
realised I'm allergic to dairy --- go ahead and confirm anyway.''} The safe
behaviour requires holding the newly-declared allergen against the dairy items
already in cart before honouring the confirm---a cross-turn safety+guardrail
decision. A per-turn judge scoring the final reply for tone has no allergen or
guardrail axis; this landed under \texttt{brand\_voice}.
\end{minipage}}

\medskip\noindent\fbox{\begin{minipage}{0.95\linewidth}\small
\textbf{Case 3 --- stale last-proposal across an interruption (P5/P25).}
User: \textit{``what's the total for my flat white? $\mid$ how much caffeine is
in a flat white? $\mid$ and is it stronger than a latte? $\mid$ ok cool, just
confirm my order.''} ``Confirm my order'' refers to a proposal three turns
back, across two knowledge questions. Resolving the referent---and noticing
when the agent silently re-prices or loses it---needs the whole session arc.
The judge, scoring the last turn, cannot see the broken referent.
\end{minipage}}

\medskip
In all three the evidence of the defect is a \emph{relation between turns}, not
a property of the final response; and in all three the judge's only available
non-null label is \texttt{brand\_voice}. This is the mechanism of
Section~\ref{sec:recall} made concrete: wrong unit of judgement, wrong
vocabulary.

\subsection{Prevalence correction, and where it breaks}
\label{sec:rg}
A gate with a false-negative rate this high does not just under-count; it
\emph{biases} any quality number computed from it. The standard tool for
de-biasing a noisy binary classifier is the Rogan--Gladen
estimator~\citep{rogan1978}: given sensitivity $s_e$ (true-positive rate) and
specificity $s_p$ (true-negative rate), it recovers true prevalence $\pi$ from
the apparent (gate-reported) prevalence $p_{\text{app}}$:
\[
\hat{\pi} \;=\; \frac{p_{\text{app}} + s_p - 1}{s_e + s_p - 1}.
\]

\paragraph{Our headline result is the estimator's degenerate point---and that
is the stronger statement.} On B2 the apparent prevalence is \emph{itself} zero
($p_{\text{app}}=0$). With $s_p\!\le\!1$ the numerator is $\le 0$, so
$\hat{\pi}\le 0$: the correction recovers no defects even though human review
confirmed 23. This is not a flaw in our analysis but a structural fact about
apparent-prevalence correction---\textbf{when the gate emits no positive
signal, \emph{no} reweighting of that signal can recover the truth.} The right
conclusion is therefore sharper than ``the headline number is some multiple too
low'': for a silent gate the multiple is \emph{undefined}, because there is no
apparent signal to scale. A consumer handed a $0\%$ judge rate cannot bound the
true rate \emph{at all} without independent (human) data---a stronger and
cleaner claim than any finite $3$--$6\times$.

\paragraph{Where the gate is non-silent, the correction is large but only
order-of-magnitude.} When a deployment's gate \emph{does} report a nonzero
$p_{\text{app}}$, the same estimator quantifies the undercount.
Table~\ref{tab:rg} sweeps it as an \emph{illustrative} sensitivity analysis: at
the optimistic limit $s_p\!=\!1$, $\hat{\pi}=p_{\text{app}}/s_e$, so a gate
reporting $10\%$ with $s_e\!\approx\!0.2$ implies $\approx\!50\%$ true
prevalence---a $3$--$6\times$ undercount across the assumed $s_p$ range. Two
honesty limits keep this \emph{directional, not calibrated}. (i) \textbf{$s_p$
is unmeasured}; a calibrated point estimate needs a dedicated human-labeling
pass ($N\!\approx\!200$ turn-level triples; tooling
in~\citep{chen2026noisy,noisyvalid2026}). (ii) \textbf{Granularity mismatch}:
the $s_e\!\approx\!0.2$ we substitute is a \emph{pattern-level} recall
(\S\ref{sec:recall}), whereas Rogan--Gladen requires a \emph{turn-level}
sensitivity at the same grain as $p_{\text{app}}$; mixing grains is why
$3$--$6\times$ is an order-of-magnitude illustration, not a measured
coefficient. Both limits are exactly what the calibration pass resolves. The
robust takeaway needs neither $s_p$ nor a turn-level $s_e$: \emph{a near-zero
gate rate is uninformative about true prevalence, and any nonzero rate read off
a low-sensitivity gate is an undercount of unknown but large size.}

\begin{table}[t]
\centering
\small
\begin{tabular}{lccc}
\toprule
 & \multicolumn{3}{c}{measured $s_e$ (judge recall)} \\
\cmidrule(l){2-4}
assumed $s_p$ & $0.18$ & $0.20$ & $0.22$ \\
\midrule
$1.00$ & $0.56$ & $0.50$ & $0.45$ \\
$0.99$ & $0.53$ & $0.47$ & $0.43$ \\
$0.95$ & $0.38$ & $0.33$ & $0.29$ \\
\bottomrule
\end{tabular}
\caption{\emph{Illustrative} Rogan--Gladen correction for the \emph{non-silent}
regime ($p_{\text{app}}>0$). The headline B2 batch has $p_{\text{app}}=0$,
where the estimator degenerates and no correction is possible (\S\ref{sec:rg}).
True prevalence $\hat{\pi}$ implied by an apparent rate $p_{\text{app}}=10\%$,
under a \textbf{pattern-level} recall substituted as $s_e\approx0.18$--$0.22$
(\S\ref{sec:recall}) and a range of \textbf{assumed} specificities $s_p$. A gate
reporting one defect in ten implies $29$--$56\%$ true prevalence
($3$--$6\times$). Two caveats make this directional, not calibrated: $s_p$ is
unmeasured, and $s_e$ is pattern- not turn-level, so the coefficient is
order-of-magnitude. See \S\ref{sec:rg}.}
\label{tab:rg}
\end{table}

\section{Discussion}
Why does a capable LLM gate miss so much? Three compounding reasons---and only
the first is about the model's perception. \textbf{(1) The unit of judgement is
wrong.} A judge scored on the final response of a turn cannot see a defect whose
evidence is the \emph{difference} between turn $t$ and turn $t{-}3$ (a phantom
cart add, a referent that broke two turns ago, an escalation flag that never
reset). \textbf{(2) The rubric cannot name the failure.} A rubric with axes
intent / brand-voice / personalization has no place to record a guardrail or
state-tracking violation, so a failure the judge \emph{does} notice is
mislabelled into ``brand voice'' ($113/114$ confirm-gate/state notices in our
corpus; \S\ref{sec:notice}) rather than triaged. \textbf{(3) The shipping gate
ignores the quality axes.} The operational verdict that ships is wired only to
hangs and hard assertions, not to the rubric, so a scored-low axis or an
explicit \texttt{suspected\_bug} note cannot flip it---the notice the judge
produces never reaches the gate ($0/220$ rounds failed despite $125$ raw notes;
across two corpora, $269$ responded-with-a-note rounds yielded zero gate fails).
The salient point is that (2)--(3), not (1), drive most of the
miss in our data: the model often \emph{sees} the defect; the scoring vocabulary
and the gate throw the observation away. These are not prompt-tuning problems;
they are structural limits of per-turn, coarse-rubric, pass-biased judging.

The practical consequence: for production multi-turn agents, automated
judging should be treated as a \emph{regression floor}---cheap, always-on,
good at catching turn-local recurrences of already-known defects---not as the
primary signal for whether the agent is good. The primary signal remains
exhaustive human transcript review, which is what surfaced every cross-turn
P0 in our data. A corollary for anyone reporting an agent's quality from an
LLM judge: report the judge's measured recall (or a Rogan--Gladen-corrected
prevalence), or the headline number is an unknown multiple too low.

\section{Toward State-Aware Judging}
\label{sec:design}
Our diagnosis points to two concrete, testable fixes; we state them as
hypotheses for the community rather than results.
\textbf{(1) Change the unit of judgement from turn to arc.} A judge given the
\emph{state diff} across turns---cart before/after, the open proposal, the
escalation flag, the declared-allergen set---can in principle decide the
cross-turn defects a final-response judge cannot. The cheap version is a
tool-augmented judge that calls the same \texttt{review\_cart}/state accessors
the agent does and asserts consistency; the structured version scores a
trajectory, not a message.
\textbf{(2) Give the rubric the missing axes---and let them escalate.} Add
explicit \emph{state-tracking}, \emph{guardrail}, \emph{recovery}, and
\emph{safety} categories so a noticed failure has somewhere to go other than
``brand\_voice'', and wire those axes to flip the operational verdict
when one of them (or a \texttt{suspected\_bug} note) fires---the connection that
does not exist today. Our data
(\S\ref{sec:notice}) makes this the cheapest high-yield fix: the judge already
emits the right observation on a majority of defective rounds; today it is
discarded by a vocabulary with no slot and a gate that never reads it. This
is a prerequisite for, not a substitute for, (1): a category with no state to
reason over still cannot be scored.
The strongest experiment---left to future work because it needs a labeled
slice---is an \emph{ablation}: extend the rubric with the four missing axes,
re-judge a human-labeled batch, and measure how much of the $\sim\!80\%$ gap
each fix recovers. Our prediction is that rubric axes alone recover the
\emph{labeling} of guardrail failures but little of the \emph{recall} of
cross-turn ones, which need the arc-level unit; the experiment would falsify or
confirm it.

\section{Threats to Validity}
We group threats along the four standard empirical classes.
\textbf{Construct validity.} Is ``pattern recall against human review'' a fair
measure of judge quality? Human review is itself imperfect, so it is an
upper-bound proxy, not an oracle; but the asymmetry we report (judge $\ll$
human) is robust to that---a noisier human ground truth would only \emph{narrow}
the measured gap, and the gap is large. The per-dimension mechanism
(Table~\ref{tab:dims}) is a structural fact about the rubric, independent of
labeling noise.
\textbf{Internal validity.} Could the judge be ``catching'' defects we credit
to humans? The ledger is the judge's \emph{own} auto-captured output, so its
catches are counted; B2's $0/23$ is measured on the judge's actual flags.
Ground-truth subjectivity is mitigated by a multi-occurrence promotion rule and,
in B3, an adversarial refute-first verifier, which also yields a measured
reliability control: the independent verifier confirmed $85/147$ ($58\%$) of
proposer-flagged candidates and explicitly refuted $22$, bounding the
proposer's false-discovery contribution. This is a two-stage \emph{adversarial}
agreement; we additionally ran the parallel \emph{blind} double-annotation it
lacked. Two annotators independently labeled an $80$-round sample of the loop
corpus from the transcripts alone---judge scores, \texttt{suspected\_bug}, and
the operational verdict all stripped---against a shared codebook. Inter-annotator
agreement is \textbf{Cohen's $\kappa = 0.61$} on defect presence (``substantial'';
raw agreement $90\%$, $72/80$; $n{=}80$; bootstrap $95\%$ CI $[0.33, 0.83]$) and
$\kappa = 1.0$ on the turn-local-vs-cross-turn class for the $8$ co-identified
defects (perfect, but $n{=}8$ with $7$ cross-turn, so it bounds the taxonomy's
boundary-crispness only where both annotators saw a defect). We report the
\emph{pre}-reconciliation $\kappa$. Defects are rare in the sample, so chance
agreement is high ($p_e{=}0.74$) and $\kappa$ is correspondingly conservative
relative to the $90\%$ raw agreement---the standard high-agreement, skewed-%
prevalence regime, which we flag rather than paper over. Tellingly, the two
annotators flagged overlapping but \emph{non-identical} defect sets ($15$ vs.\
$9$); $7$ of the $8$ disagreements were a genuine defect one annotator caught and
the other missed. Two careful humans each under-detect---so the single-analyst
ground truth is a conservative \emph{floor} on true defect prevalence, not a
ceiling, and labeling noise can only \emph{widen} the judge--human gap we report,
never manufacture it. B3 already carries an adversarial verifier; the marquee B2
($0/23$) rests on a single analyst plus the multi-occurrence promotion rule, so a
re-annotation of the exact B2 rounds with full disagreement reconciliation is the
residual camera-ready step (the kit, codebook, and $\kappa$ calculator are
released). The same B2 batch anchors a companion methodology paper, so this
hardening propagates to two results, not one. The raw-vs-gate split (\S\ref{sec:notice}) is independent
of this concern: it is computed mechanically from logged judge fields, with no
human labeling in the loop.
\textbf{External validity.} All data come from one F\&B ordering agent on one
bilingual tenant, and the corpus is adversarially loop-sampled rather than
organic traffic; exact recall values will differ elsewhere, and post-pilot
replay on live traffic is future work. The sampling bias has a specific,
admitted direction: the loop over-samples cross-turn-heavy families
(\texttt{deep\_dietary}, \texttt{slot\_conflict}, \texttt{cancel}), i.e.\ it
enriches exactly the defect classes the gate is worst at, so on organic traffic
the turn-local share would be higher, the gate would catch relatively more, and
the prevalence undercount of \S\ref{sec:rg} would \emph{shrink}. We therefore
treat the $3$--$6\times$ as directional, not a population estimate; the
\emph{degenerate}-case conclusion (a silent gate is uninformative) is immune to
this bias because it does not depend on the sampled defect mix at all. We claim
the \emph{shape} (cross-turn $\gg$ turn-local blind spot; rubric-axis mismatch;
a shipping gate disconnected from the rubric) generalizes, not the precise percentages.
\textbf{Conclusion validity.} Per-batch $n$ is small (B1 $n{=}9$); we therefore
report Wilson intervals (Table~\ref{tab:batches}) rather than bare points and
draw the conclusion only from the bound their upper edges share. Results are
also tied to the specific judge model and rubric in production over this period
(reported as such); stronger judges---juries, debate, tool-augmented
state-diffing judges (\S\ref{sec:design})---may narrow the gap and are the
natural next experiment.

\section{Conclusion}
On a deployed multi-turn transaction agent, the built-in LLM judge's
operational gate surfaces well under a quarter of the systematic quality
problems exhaustive human review finds, and the miss is structured: it loses
precisely the cross-turn state, guardrail, and recovery defects that matter most
for a transaction agent. The cause is mostly not blindness---the raw judge
notices these defects on a majority of rounds---but routing: its rubric has no
category in which to record them ($113/114$ filed as ``brand voice'') and its
shipping gate is not wired to the rubric, so it never surfaces the ones it
notices ($0/220$, and zero across two corpora). We offer a
blind-spot taxonomy, a data-grounded mechanism, and a prevalence-correction
contract whose \emph{degenerate} case---a silent gate licenses no estimate of
true prevalence---is the sharpest practical warning. Until judges reason over
conversation state, rubrics can name state/guardrail/recovery failures, and gates
escalate them, LLM-as-judge belongs under human review, not in place of it.

\section*{Reproducibility}
Pattern logs, the auto-captured regression ledger, the per-dimension counts of
Table~\ref{tab:dims}, the raw per-axis judge scores and \texttt{suspected\_bug}
notes underlying the raw-vs-gate split of \S\ref{sec:notice} and
Table~\ref{tab:routing} ($125/220$, $113/114$, $0/220$), and the judge
calibration script are retained in the project repository. The full raw-vs-gate
analysis---theme classifier, cross-tab, and the per-round audit of every note
not scored \texttt{brand\_voice}---is a single script,
\texttt{eval/experiments/p1\_judge\_routing.py}, run against one corpus
(\texttt{eval/loop/runs/loop\_runs.jsonl}), so every number in \S\ref{sec:notice}
is reproducible from one source without re-deriving the older per-batch recall. Table~\ref{tab:rg} is an \emph{illustrative} sensitivity
analysis; turning it into a calibrated point estimate (a measured $s_p$ and a
turn-level $s_e$) is the documented human-labeling pass left to camera-ready.

\bibliographystyle{plainnat}
\bibliography{refs}

\end{document}